\documentclass[11pt,a4paper]{article}
\usepackage[hyperref]{acl2021}
\usepackage{times}
\usepackage{latexsym}

\usepackage{microtype}
\usepackage{times}
\usepackage{epsfig}
\usepackage{graphicx}
\usepackage{amsmath}
\usepackage{amssymb}
\usepackage{kotex}
\usepackage{multirow}
\usepackage{makecell}

\usepackage{mathtools}
\usepackage{color}

\aclfinalcopy % Uncomment this line for the final submission
\newcommand{\R}{\mathbb{R}}

\title{Learning Dynamic BERT via \\ Trainable Gate Variables and a Bi-modal Regularizer}
\author{Seohyeong Jeong \\
  Seoul National University \\
  \texttt{seo\_hyeong@snu.ac.kr} \\\And
  Nojun Kwak \\
  Seoul National University \\
  \texttt{nojunk@snu.ac.kr} \\}

\date{}

\begin{document}
\maketitle
\begin{abstract}
The BERT model has shown significant success on various natural language processing tasks. However, due to the heavy model size and high computational cost, the model suffers from high latency, which is fatal to its deployments on resource-limited devices. To tackle this problem, we propose a dynamic inference method on BERT via trainable gate variables applied on input tokens and a regularizer that has a \textit{bi-modal} property. Our method shows reduced computational cost on the GLUE dataset with a minimal performance drop. Moreover, the model adjusts with a trade-off between performance and computational cost with the user-specified hyperparameter.  
\end{abstract}

\section{Introduction}
BERT \cite{devlin2018bert}, the large-scale pre-trained language model, has shown significant improvements in natural language processing tasks \cite{dai2015semi, rajpurkar2016squad, mccann2017learned, peters2018deep,  howard2018universal}.
%Furthermore, \citet{gururangan2020don} has shown that can these models can enjoy performance gains with further pre-training on the additional corpus. 
However, the model suffers from the heavy model size and high computational cost, which hinders the model to be applicable in real-time scenarios on resource-limited devices. 
% BERT[] and RoBERTa[] are transformers[] based large-scale pre-trained language models and have brought groundbreaking improvements to NLP tasks. Furthermore, [Don't stop pretraining] has shown that can these models can enjoy performance gains with further pre-training on additional pre-training corpus.
% However, models suffer from heavy model sizes and high computational cost, which hinders the model to be applicable in real-time scenarios on resource-limited devices.
\begin{figure}[t]
\begin{center}
\includegraphics[width=0.9\linewidth]{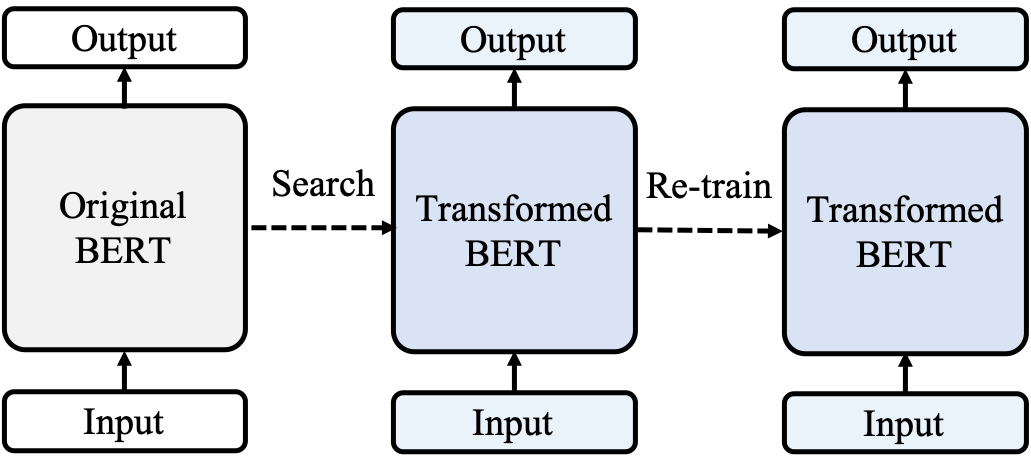}\\
(a) Two-stage method framework\\
\vspace{0.3cm}
\includegraphics[width=0.8\linewidth]{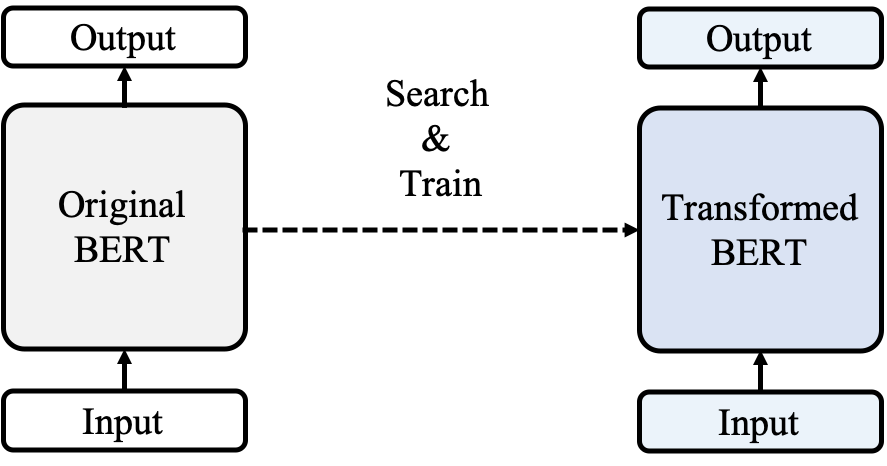}\\
(b) One-stage method framework\\
\end{center}
   \caption{The overview of the two-stage method and the one-stage method for dynamic inference models. 
   %Our method falls in the category of (b).
   }
\label{fig:twostage_onestage}
\end{figure}

\begin{figure*}[t]
\begin{center}
  \includegraphics[width=0.9\linewidth]{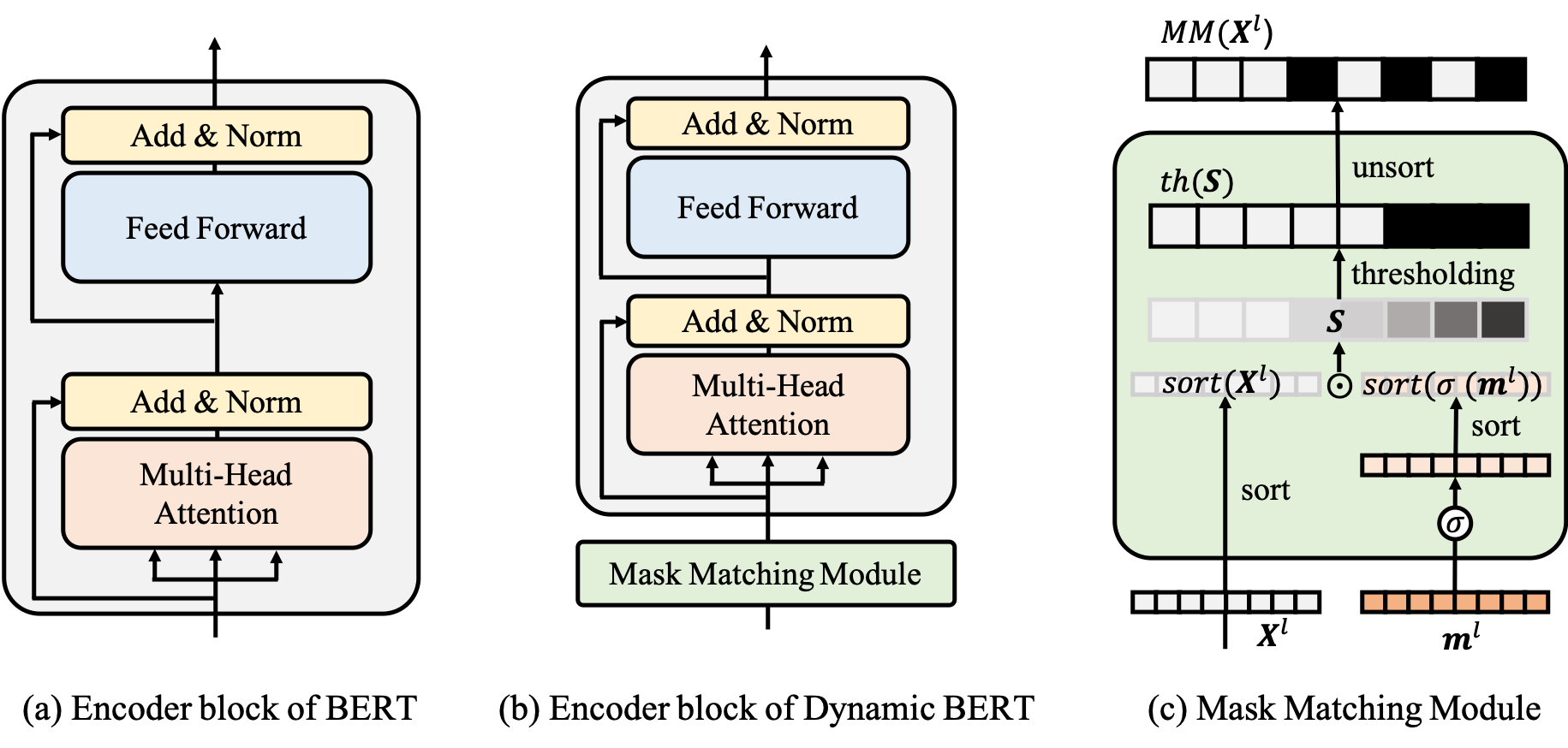}
\end{center}
 \caption{Comparison of an original encoder block of BERT and our model with the mask matching module.}
\label{fig:mmm}
\end{figure*}

\citet{khetan2020schubert} has shown that using a ``tall and narrow'' architecture provides better performance than a ``wide and shallow'' architecture when obtaining a computationally lighter model. Inspired by this finding, we propose a task-agnostic method to dynamically learns masks on the input word vectors for each BERT layer during fine-tuning. To do this, we propose a gating module, which we call a mask matching module, that sorts and matches each of the input tokens to corresponding learned masks. Note that we use the word ``gate'' and ``mask'' interchangeably. Inspired by \citet{srinivas2017training}, we train our model with an additional regularizer that has $bi$-$modal$ property on top of the $l_{1}$-variant regularizer, that we suggest in this work, and an original task loss. Using a $bi$-$modal$ regularizer allows the model to learn a downstream task and search the model architecture simultaneously, without requiring any further fine-tuning stage.

In this paper, we conduct experiments with BERT-base on the GLUE \cite{wang2018glue} dataset and show that the mask matching module and the $bi$-$modal$ regularizer enable the model to search the architecture and fine-tune on a downstream dataset simultaneously. Compared to previous works in compressing or accelerating the inference time of BERT \cite{sanh2019distilbert, jiao2019tinybert, sun2019patient, liu2020fastbert}, our method possesses three main differences. First of all, our method allows task-agnostic dynamic inference rather than a single reduced-sized model. Secondly, our method does not require any additional stage of fine-tuning or knowledge distillation (KD) \cite{hinton2015distilling} and lastly, our method provides a hyperparameter that can be specified by a user to control a trade-off between the computational complexity and the performance.

\section{Related Work}

% layer dropping: poor man's BERT, DeeBERT
% sequence dropping: PowerBERT
% hybrid: dynaBERT
% dynamic inference: DeeBERT, FastBERT, DynaBERT (dynamic size both)
% DeeBERT (dynamic early exiting), FastBERT: layer dropping, double stage of either pretraining the backbone using KD or finetuning after the structural achictecture search

There have been numerous works to compress and accelerate the inference of BERT. Adopting KD, \citet{sanh2019distilbert} attempts to distill heavy teacher models into a lighter student model. Pruning the pre-trained model is another method to handle the issue of heavy model sizes and high computational cost \cite{michel2019sixteen, gordon2020compressing}. \citet{sajjad2020poor} prunes BERT by dropping unnecessary blocks and \citet{goyal2020power} does it by dropping semantically redundant word vectors. Some other works have introduced dynamic inference to accelerate the inference speed on BERT. \citet{xin2020deebert} allows early exiting and \citet{liu2020fastbert} adjusts the number of executed blocks dynamically.
% Other approaches of model compression, such as quantization [] and parameter sharing [ALBERT] have been explored as well. 

Our work is mainly inspired by \citet{goyal2020power} and we integrate dynamic inference to the sequence pruning. The main difference of our method compared to theirs and other pruning methods \cite{michel2019sixteen, gordon2020compressing, sajjad2020poor} is that our model allows a task-agnostic dynamic inference without the additional requirement of fine-tuning after a model architecture search, as illustrated in Figure \ref{fig:twostage_onestage} (b). There exist other works \citet{hou2020dynabert, goyal2020power, liu2020fastbert, fan2019reducing, elbayad2019depth} to dynamically adjust the size and latency of the language models. However, these approaches either works in a two-stage setting where further fine-tuning or knowledge distillation is required, as shown in Figure \ref{fig:twostage_onestage} (a) or consider a depth-wise compression rather than a width-wise compression. We experimentally show that the computational cost can be reduced with minimal performance drop on GLUE \cite{wang2018glue}.

\section{Method}
In this section, we introduce our proposed method that mainly consists of a mask matching module and an additional regularizer to induce polarization on mask variables.

\subsection{Mask Matching Module}
\label{sec:mmm}
As presented in Figure \ref{fig:mmm} (a), the original encoder block of BERT consists of multi-head attention and feed-forward networks. 
% should this be in introduction?
The intuition behind the mask matching module is to filter out input tokens that do not contribute as much in solving a given task so that the model can benefit from the reduced computational burden during the process of multi-head attention. Since the multi-head attention on sequences of length, $L$ is $O(L^{2})$ in computational complexity, we expect to reduce this cost by masking out unnecessary tokens for each encoder block. 

In order to learn important tokens in the training process, we introduce the mask matching module which is placed before the original encoder block of BERT, as shown in Figure \ref{fig:mmm} (b). Figure \ref{fig:mmm} (c) shows the detailed process of the mask matching module. The superscript $l$ represents the $l^{th}$ block, which we omit from the following description in this section. The module consists of sorting input tokens according to importance scores and matching each input token to a mask, and thresholding the computed tokens with a certain value.

We first compute the importance score, $\textbf{s}\in \R^{I}$, of each token in the input sequence as $s_i = \sum_{j=1}^{J}|\textbf{X}_{ij}|$, where $\textbf{X} \in [I, J]$ is the matrix representation of the input, with $I$ being the length of the input sequence and $J$ being the size of the hidden dimension. As each token has corresponding $s_i$, we sort the input matrix, $\textbf{X}$, input sequence-wise according to the importance score of each token. Then, sorted input matrix and sorted masks are multiplied element-wise to perform mask matching to obtain a masked matched input matrix, $\textbf{S} \in [I, J]$:
\begin{equation}
\label{eq:mask_matching}
\textbf{S} = sort(\textbf{X}) \odot expand(sort(\sigma(\textbf{m})))
\end{equation}
where $\sigma()$ is a sigmoid function and $\textbf{m} \in \R^{I}$ is a parameter. Note that since $\sigma(\textbf{m}) \in \R^{I}$ and $\textbf{X} \in [I, J]$, we \textit{expand} $\sigma(\textbf{m})$ to match the shape of $\textbf{X}$ by multiplying it $J$ times and stacking them.

Then, we introduce a thresholding scheme on masked tokens as follows:
\begin{equation}
\label{eq:thresholding}
th(\textbf{S}_{i,1:J}) = 
\begin{cases}
    \textbf{S}_{i,1:J} & \text{$\sigma(m_i) \geq \alpha$} \\
    \textbf{0} & \text{$\sigma(m_i)< \alpha$}
\end{cases}
\end{equation}
where $\alpha$ is a hyperparamter and $m_i$ is a learned mask value for $i^{th}$ token in the input. The thresholed output is unsorted into the original sequence of input tokens and passed to the consecutive encoder block as an input. The final output of the masked matching module is written as follows:
\begin{equation}
\label{eq:final_mask_matching}
\textbf{X}^{m} = unsort(th(\textbf{S}))
\end{equation}

\subsection{Inducing Polarization}
\label{sec:polar}
% Previous works \cite{goyal2020power} \sh{find other works} have explored inducing sparsity to neural networks with $l_1$ or $l_2$ regularizers. However, these regularizers do not guarantee well-polarized values for a gate(mask) variable. 
Traditional $l_1$ and $l_2$ regularizers do not guarantee well-polarized values for a gate(mask) variable. In order to induce polarization on our masks, we utilize a \textit{bi-modal} regularizer proposed by \cite{murray2010algorithm, srinivas2017training} to learn binary values for parameters. \cite{srinivas2017training} used an overall regularizer which is a combination of the $bi-modal$ regularizer and a traditional $l_1$ or $l_2$ regularizer. In this work, we use a customized regularizer, which is a variant of $l_{1}$, denoted as $l_{filter}$, to dynamically adjust the level of sparsity according to the user-specified hyperparameter. 
\begin{equation}
\label{eq:l_filter_1}
    \begin{gathered}
    l_{filter} = \frac{1}{L} \sum_{n=1}^{L}|\textbf{v}_{filter, l}|, \\
    \textbf{v}_{filter} = \textbf{w} \odot (\textbf{v}_{masks} - \textbf{v}_{user}).
    \end{gathered}
\end{equation}
$\textbf{v}_{masks}$, $\textbf{v}_{user}, $\textbf{w}$ \in \R^{L}$ are filtering weights, mass of masks, and the user specified mass of masks with $L$ being the number of blocks in a model.
\begin{equation}
\label{eq:l_filter_2}
\begin{gathered}
    \textbf{v}_{masks, l}=\sum_{i=1}^{I}\sigma(m_{i}^{l}), \\
    \textbf{v}_{user, l}=I \times L \times \gamma, \\
    \textbf{w}_{l}=1.5- \{\sum_{i=1}^{I} \sigma(m_{i}^{l})\}/I.
    \end{gathered}
\end{equation}
where $I$ is the length of the input token sequence and $0 \leq \gamma \leq 1$ is a hyperparameter to enforce the user-specified level of filtering tokens in the model. 
Then, the polarization regularizer is written as a linear combination of $l_{filter}$ and $l_{bi-modal}$, which has a form of $w\times(1-w)$, as follows:
\begin{equation}
\begin{aligned}
L_{polar} & = \lambda_{filter} * l_{filter} \\
& + \lambda_{bi} * \sum_{l=1}^{L}\sum_{i=1}^{I} \sigma(m_{i}^{l}) (1-\sigma(m_{i}^{l})).
\end{aligned}
\end{equation}

Our total objective function is stated as follows:
\begin{equation}
\label{eq:total_loss}
L_{total} = L_{task} + L_{polar}
\end{equation}
$L_{task}$ is the loss for a downstream task. We show the effect of the \textit{bi-modal} regularizer in Sec. \ref{sec:ablation}.

\begin{table*}[t]
\centering
\resizebox{1.0\textwidth}{!}{
\begin{tabular}{l | c c c c c c c c}
\hline\hline
\multirow{2}{*}{\textbf{Models}}  & \multicolumn{8}{c}{\textbf{GLUE-test}} \\

 & MNLI-(m/mm)  & QNLI  & QQP  & RTE & SST-2  & MRPC & CoLA & STS-B  \\  \hline
BERT-base \cite{devlin2018bert} &84.6 / 83.4  & 90.5  & 71.2  & 66.4 & 93.5 & 88.9  & 52.1 & 85.8  \\ 
BERT-base-ours  & 84.5 / 83.7  & 90.7  &71.8   &62.4  & 93.9 & 83.7  & 51.2 & 78.9 \\ 
(FLOPs) & 10872M  & 10872M  & 10872M   &10872M  & 10872M & 10872M  & 10872M & 10872M \\

\hline
\textbf{Ours $(\gamma=0.3)$} & \textbf{82.2} / \textbf{81.8} & \textbf{87.5}  & \textbf{69.9} & \textbf{58.2} & \textbf{92.8} & \textbf{84.9} & \textbf{33.3} & \textbf{79.5} \\
\multirow{2}{*}{(FLOPs)} 
& 3357M     & 3915M     & 3766M  & 4629M    & 3887M & 4417M  & 2629M & 3371M  \\
& \textbf{(3.23$\times$)}  & \textbf{(2.77$\times$)}  & \textbf{(2.88$\times$)}  & \textbf{(2.35$\times$)}  & \textbf{(2.80$\times$)} & \textbf{(2.46$\times$)}  & \textbf{(4.13$\times$)} & \textbf{(3.23$\times$)} \\

\hline \hline
\end{tabular}
}
\caption{Comparison of GLUE test results, scored by the official evaluation server. BERT-ours is our implementation of the baseline model, BERT. Performances for \textbf{Ours} is reported with $\gamma=0.3$. 
%QQP and MRPC are reported with F1 scores, STS-B is reported with Spearman correlations and other tasks are reported with accuracy. 
Last row shows the computational improvement compared to the FLOPs of original BERT-base.}
\label{tab:perf_glue}
\end{table*}

\begin{table}[t]
\centering
\resizebox{\linewidth}{!}{
\begin{tabular}{l | c c c c c}
\hline\hline
\multirow{2}{*}{\textbf{Models}}  & \multicolumn{3}{c}{\textbf{GLUE-eval}} \\

                    & MNLI-(m/mm)   & QNLI      & SST-2  \\  \hline
                    
BERT-base-ours      &84.3 / 84.9    & 91.7      & 92.5  \\ 
(FLOPs)             & 10872M        & 10872M    & 10872M \\
\hline
Ours $(\gamma=0.1)$ &70.8 / 71.0    & 73.6      & 87.0  \\ 
(FLOPs)             & 1907M        & 1872M    & 1865M   \\
\hline
Ours $(\gamma=0.2)$ &79.0 / 79.0    & 85.5      & 91.3  \\ 
(FLOPs)             & 2904M        & 2883M    & 2883M   \\
\hline
\textbf{Ours $(\gamma=0.3)$} &82.4 / 82.6    & 88.7      & 91.6  \\ 
(FLOPs)             & 3357M        & 3915M    & 3887M   \\
\hline
Ours $(\gamma=0.4)$ &82.9 / 83.7    & 89.8      & 92.2  \\ 
(FLOPs)             & 4919M        & 4926M    & 4863M   \\
\hline
Ours $(\gamma=0.5)$ &83.1 / 83.7    & 90.4      & 91.6  \\ 
(FLOPs)             & 5923M        & 5994M    & 5916M   \\
\hline
Ours $(\gamma=0.6)$ &83.1 / 83.7    & 90.6      & 91.6  \\ 
(FLOPs)             & 6962M        & 7033M    & 6676M   \\
\hline
Ours $(\gamma=0.7)$ &83.2 / 83.9    & 89.8      & 92.0  \\ 
(FLOPs)             & 9218M        & 9027M    & 9182M   \\
\hline
Ours $(\gamma=0.8)$ &83.8 / 84.2    & 89.8      & 92.2  \\ 
(FLOPs)             & 10829M        & 10398M    & 9818M   \\
\hline
Ours $(\gamma=0.9)$ &83.9 / 84.4    & 91.1      & 92.4  \\ 
(FLOPs)             & 10872M        & 10872M    & 10872M   \\

\hline \hline
\end{tabular}
}
\caption{Performances and FLOPs on GLUE evaluation set with different values of $\gamma$.}
\label{tab:perf_glue_gamma}
\end{table}

\begin{table}[t]
\centering
\resizebox{\linewidth}{!}{
\begin{tabular}{l | c c c c c}
\hline\hline
\multirow{2}{*}{\textbf{Models}}  & \multicolumn{3}{c}{\textbf{GLUE-eval}} \\

                    & MNLI-(m/mm)   & QNLI      & SST-2  \\  \hline
                    
Ours $(\lambda_{bi}=2.0)$ &82.4 / 82.6    & 88.7      & 91.6  \\ 
Ours $(\lambda_{bi}=0.0)$ &60.9 / 61.3    & 67.2      & 76.0  \\ 

\hline \hline
\end{tabular}
}
\caption{Ablation study of the \textit{bi-modal} regularizer on the GLUE evaluation set.}
\label{tab:abla}
\end{table}

\section{Experiments}
We evaluate the proposed method on eight datasets in GLUE \cite{wang2018glue} benchmark.

\subsection{Implementation Details}
We fine-tune the pre-trained BERT-base model on 8 datasets in the GLUE benchmark dataset for 3 epochs with a batch size of 128. The hidden dimension is set to $J=768$ and the length of the input token sequence is set to $I=128$. For the rest of the details, we follow the original settings of BERT.

We set $\alpha=0.5$, $\lambda_{filter}=0.01$, and $\lambda_{bi}=2.0$. We use the separate Adam \cite{kingma2014adam} optimizer for training mask variables. The Adam optimizer for mask variables are set with initial learning rate of $0.05$ with two momentum parameters $\beta_{1}=0.9$ and $\beta_{2}=0.999$, and $\epsilon=1\times 10^{-8}$. Mask variables are initialized with random values from a uniform distribution on the interval [0, 1). We do not introduce the mask variables for the very first block in the model. Additionally, we never filter (do not mask) the first token of each input, the special token \texttt{[CLS]}. Introducing mask variables results in ``the length of tokens $\times$ the number of blocks'' additional number of parameters.

\subsection{Main Results}
% mention specific number??
We compare our model with the BERT-base baseline. Table \ref{tab:perf_glue} summarizes the results of these models. Performances on the first row are taken from \citet{devlin2018bert} and we show performances with our implementation on the second row. The last row shows the improvement compared to the FLOPs of the baseline model. It shows that our dynamic inference method with $\gamma=0.3$ shows minimal degradation on GLUE datasets with an average of 3 times fewer FLOPs. Furthermore, our model works in a task-agnostic manner and outputs the optimal architecture for each given downstream dataset, instead of a single reduced-sized model.

Table \ref{tab:perf_glue_gamma} shows that our model is capable of dynamically adjusting the computational cost with a trade-off between FLOPs and performance. It shows that the hyperparameter, $\gamma$, works properly showing proportional FLOPs to its given value. The result presents generally a consistent trade-off between FLOPs and performance.

\subsection{Ablation Studies}
\label{sec:ablation}
To analyze the effect of the \textit{bi-modal} regularizer, we conduct an ablation study by removing it from the training process. Table \ref{tab:abla} shows the effect of the \textit{bi-modal} regularizer and we claim that employing this regularizer during the training process plays a huge role in learning to perform well on a downstream task as well as searching the optimal model structure with the help of well-polarized mask variables. Further analysis on the behavior of mask variables with and without the \textit{bi-modal} regularizer is shown in Appendix \ref{sec:appendix}.

\section{Conclusions}
In this work, we explore the task-agnostic dynamic inference method on BERT that works by masking out the input sequence for each block. To do this, we propose a mask matching module and a variant of $l_{1}$ regularizer, which we call $l_{filter}$. Our method yields various levels of models with different performance and computational complexity, depending on the hyperparameter value that the user inputs. Conducting experiments on the GLUE dataset, our method shows that BERT, used with our method, can enjoy lighter computation with minimal performance degradation.

\bibliographystyle{acl_natbib}
\bibliography{acl2021}

\appendix

\begin{figure*}[t]
\begin{center}
  \includegraphics[width=1.0\linewidth]{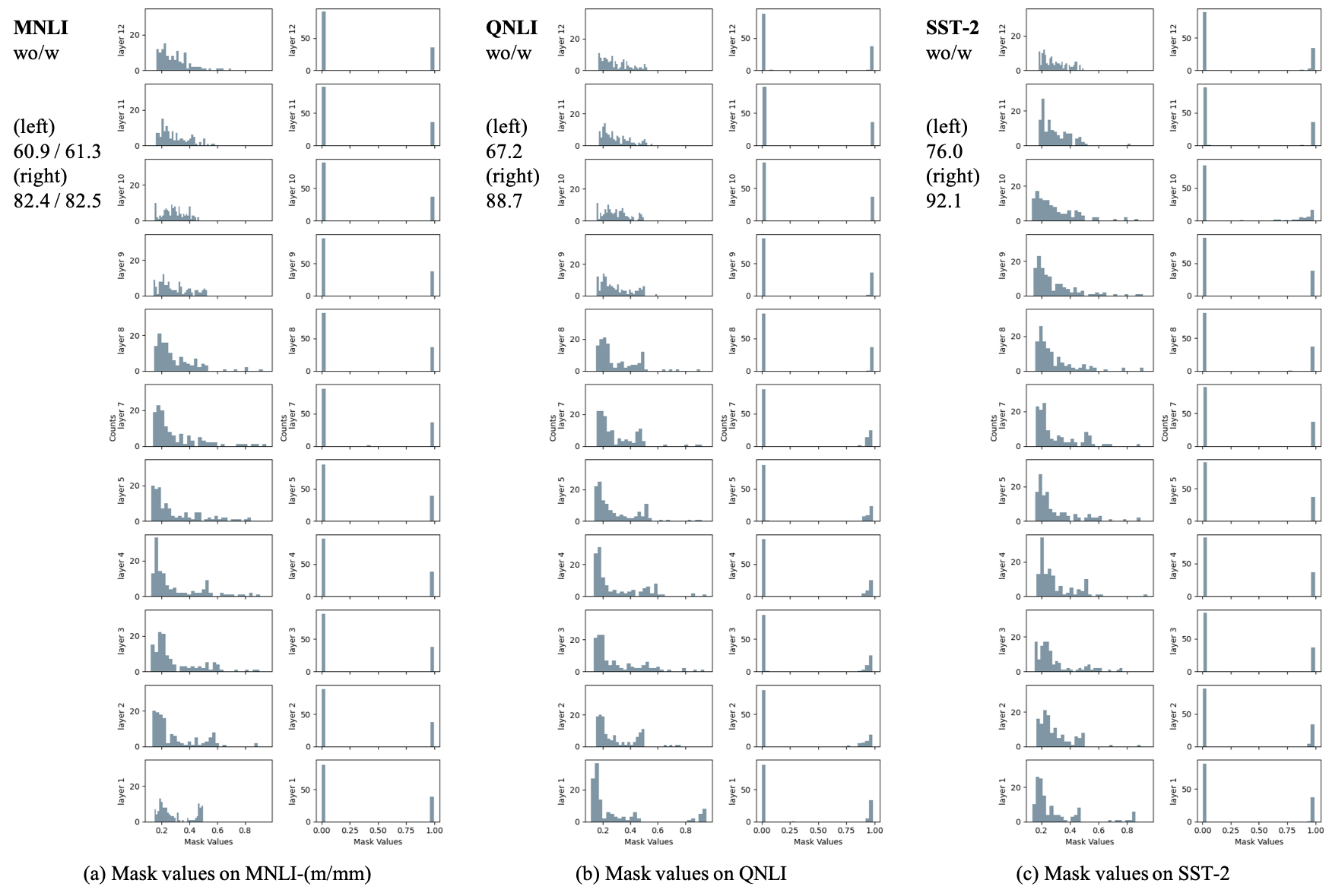}
\end{center}
 \caption{Histograms of $\sigma(\textbf{m})$ for each encoder block without and with the $l_{bi-modal}$ regularizer. For MNLI-(m/mm), QNLI, and SST-2 datasets, we show the histogram of $\sigma(\textbf{m})$ values \textbf{without} the $l_{bi-modal}$ regularizer on the left and \textbf{with} the regularizer on the right. The performance on each scenario is written on the left top corner of each figure. It shows that the $l_{bi-modal}$ regularizer not only participates in training mask variables in a well-polarized manner but also plays an important role in learning to perform well on a given task.}
\label{fig:hist_mask}
\end{figure*}

\section{Additional Details}
\subsection{Experimental Details}
An output of multi-head attention, feed-forward network, and layer normalization from Figure \ref{fig:mmm} further needs to be masked since these computations contain bias terms. Our goal is to mask out the input matrix of each encoder block token-wise. Therefore, we apply hard masking on input matrix dimensions that are masked out by the mask matching module after computations mentioned above.

\subsection{Reported Measures for GLUE}
QQP and MRPC are reported with F1 scores, STS-B is reported with Spearman correlations and other tasks are reported with accuracy. 

\section{Interpretation of $l_{filter}$ Regularizer}
We propose a variant of $l_{1}$ regularizer, called $l_{filter}$, as shown in Eq. \ref{eq:l_filter_1} and \ref{eq:l_filter_2}. As our $l_{filter}$ can come across somewhat heuristic, we explain the intuition and interpretation behind the regularizer. Let's consider an extreme case of $\sum_{i=1}^{I} \sigma(m_i^{a})=I$ and  $\sum_{i=1}^{I} \sigma(m_i^{b})=0$. Then, from the last line of Eq. \ref{eq:l_filter_2}, $\textbf{w}_a=0.5$ and $\textbf{w}_a=1.5$. This means that input tokens for the $a^{th}$ block are required more than input tokens for the $b^{th}$ block of the model, since the prior use more tokens (mask out less tokens). As shown in the second line of Eq. \ref{eq:l_filter_1}, $\textbf{w}$ works as a weight for $\textbf{v}_{masks}-\textbf{v}_{user}$. Instead of applying same weight for each block, we intend to apply weights accordingly to the number of masks used in each block. In other words, we wish to pose heavier loss on the $b^{th}$ block than the $a^{th}$ block of the model. 

\section{Analysis on Mask Variables}
\label{sec:appendix}
Figure \ref{fig:hist_mask} shows learned values for mask variables after the sigmoid function. Each histogram has mask values after the sigmoid function on the x-axis. Since we conduct experiments on BERT-base, we show results for every block in the model from the $2^{nd}$ block to the $12^{th}$ block from bottom to top in the figure. It shows that the $l_{bi-modal}$ regularizer not only participates in training mask variables in a well-polarized manner but also plays an important role in learning to perform well on a given task.

\end{document}